\documentclass[10pt,twocolumn,letterpaper]{article}

\usepackage{iccv}
\usepackage{times}
\usepackage{epsfig}
\usepackage{graphicx}
\usepackage{amsmath}
\usepackage{amssymb}

\usepackage{booktabs}
\usepackage{xcolor}
\usepackage{amsmath, bm, subfigure, epstopdf, url, pifont, overpic, cases}
\usepackage{latexsym, amssymb, bbding, multirow, makecell,  diagbox, enumitem, caption}
\usepackage{caption}
\usepackage{array, enumitem, soul, algorithm, algpseudocode}

\usepackage{arydshln}

\usepackage{multirow, multicol}
\usepackage{adjustbox}

\newlength \g

\usepackage[pagebackref=true,breaklinks=true,letterpaper=true,colorlinks,bookmarks=false]{hyperref}

\iccvfinalcopy 

\def\iccvPaperID{10} 

\begin{document}

\title{RealisMotion: Decomposed Human Motion Control and Video Generation in the World Space} 
\author{Jingyun Liang$^{1,2}$ ~ Jingkai Zhou$^{1,2,4}$ ~ Shikai Li$^{1,2}$ ~ Chenjie Cao$^{1,2}$ \\ Lei Sun$^{3}$ ~ Yichen Qian$^{1,2}$ ~ Weihua Chen$^{1,2}$ ~ Fan Wang$^{1,2}$  ~ \\
$^{1}$DAMO Academy, Alibaba Group ~~~~ $^{2}$Hupan Lab ~~~~ $^{3}$INSAIT ~~~~ $^{4}$Zhejiang University \\
{\tt\small }\url{https://jingyunliang.github.io/RealisMotion}
}


\twocolumn[{%

\maketitle

\renewcommand\twocolumn[1][]{#1}%

\begin{center}
\begin{overpic}[width=0.99\textwidth, page=1]{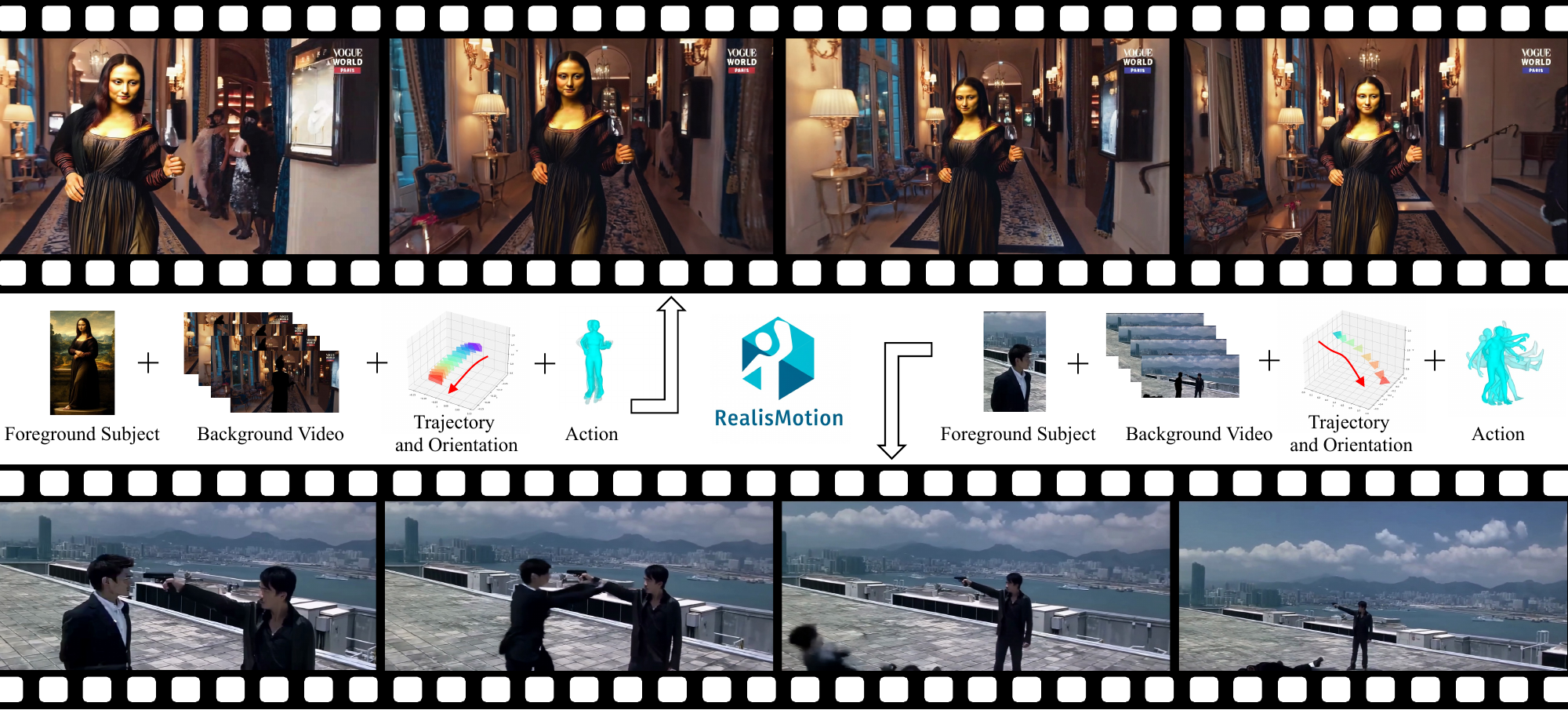}
\end{overpic}
\small
\vspace{-0.1cm}
\captionsetup{type=figure}
\caption{By decomposing the human motion into trajectory and action, and video appearance into foreground subject and background video, the proposed RealisMotion generates natural human motion videos by placing the foreground subject in the background video and having it perform the corresponding action along the specified trajectory. We provide more than 100 video examples in the project homepage \url{https://jingyunliang.github.io/RealisMotion}.}
\label{fig:teaser}
\end{center}
}]

\begin{abstract}
Generating human videos with realistic and controllable motions is a challenging task. While existing methods can generate visually compelling videos, they lack separate control over four key video elements: foreground subject, background video, human trajectory and action patterns. In this paper, we propose a decomposed human motion control and video generation framework that explicitly decouples motion from appearance, subject from background, and action from trajectory, enabling flexible mix-and-match composition of these elements. Concretely, we first build a ground-aware 3D world coordinate system and perform motion editing directly in the 3D space. Trajectory control is implemented by unprojecting edited 2D trajectories into 3D with focal-length calibration and coordinate transformation, followed by speed alignment and orientation adjustment; actions are supplied by a motion bank or generated via text-to-motion methods. Then, based on modern text-to-video diffusion transformer models, we inject the subject as tokens for full attention, concatenate the background along the channel dimension, and add motion (trajectory and action) control signals by addition. Such a design opens up the possibility for us to generate realistic videos of anyone doing anything anywhere. Extensive experiments on benchmark datasets and real-world cases demonstrate that our method achieves state-of-the-art performance on both element-wise controllability and overall video quality.
\end{abstract}

\section{Introduction}
Imagine Mona Lisa participating in a stylish event at a luxurious hotel, gracefully approaching you while holding a glass of red wine. Imagine the real cop Chan shooting the undercover police chief Lau, on a rooftop framed by the Hong Kong skyline. (See Fig.~\ref{fig:teaser} for our results.) While recent advances in human video generation and editing have shown promising results~\cite{hu2024animate, zhu2024champ, zhou2024realisdance}, existing methods still struggle to realize such creative transformations due to their limited control over individual video elements, such as subject, background, trajectory and action.

Currently, most of the existing human video generation methods are designed to transfer motions between individuals. Given a guidance video and a reference image, these methods first extract motion representations such as pose~\cite{yang2023effective, hu2024animate} and depth~\cite{hu2025animate} from the video. Then, they animate the reference image according to the extracted motion. This pipeline, whether operating in 2D image space~\cite{wang2023disco} or 3D camera space~\cite{zhu2024champ}, is limited in the following aspects. First, the foreground and background are jointly defined, which prevents independent control of the subject and the environment. Second, the tight coupling between action patterns and trajectory prevents independent manipulation of 'what' actions to perform and 'where' to perform them. Third, limited understanding of background geometry hampers editing of the subject's movement along the depth axis, making it hard to produce plausible animations with correct perspective scaling. Fourth, when the camera view changes across frames, the scene coordinate frame also shifts, complicating global trajectory control and consistent action editing. Together, these constraints lead most methods to assume that the human in both guidance video and reference image is centrally framed and near the camera, effectively reducing the task to simple motion copying.

In this paper, we introduce a decomposed human motion control and video generation framework that overcomes the limitations described above. Our key idea is to treat subject, background, trajectory, and action as independent, composable dimensions. This decomposition is realized in two stages. In the first stage, we represent human motion with the 3D parametric SMPL-X model~\cite{pavlakos2019expressive} and build a 3D world coordinate system with physical ground awareness. After freely editing the 2D image-space trajectory, we unproject it into the 3D world space using depth estimation, focal-length calibration and coordinate transformation. The moving speed and human orientation are also aligned with the real motions. Then, the corresponding action sequence is retrieved from a motion bank or synthesized with text-to-motion methods. Finally, we render depth, normal, and color maps from the 3D scene to serve as conditioning guidance for subsequent video synthesis. In the second stage, we fuse these elements into coherent videos with a video generation model based on WAN-2.1~\cite{wang2025wan}. Starting from WAN-2.1-T2V, we fine-tune the model end-to-end with three key extensions: (1) subject injection via token concatenation along the sequence dimension, (2) background incorporation by channel-wise concatenation, and (3) motion (\ie, trajectory + action) conditioning implemented with an additional ControlNet-style~\cite{zhang2023adding} module.

The contributions of this paper are summarized as follows.
\begin{enumerate}
\item We present a decomposed human motion-control and video-generation framework that models subject, background, trajectory, and action as independent, composable elements, enabling flexible mix-and-match editing. A detailed controllability comparison of related works is provided in Table~\ref{tab:traj_orient_compare}.
\item We combine 3D physical priors with a learned video diffusion prior. The physical priors handle geometry-sensitive tasks (\eg, 3D trajectory and action control, occlusion, and foreshortening) in the 3D domain, while the video diffusion prior handles appearance and temporal aspects (\eg, object/background control, frame consistency, and human–environment interaction) in the video domain.
\item We perform all trajectory and action edits in the 3D world space, preserving realistic speed, orientation, motion style and perspective effects.
\item We introduce a motion-conditioned video generation model built on the latest diffusion-transformer model Wan-2.1. Experiments on benchmark datasets and real-world cases show improved fidelity and controllability compared to prior motion-transfer methods.
\end{enumerate}

\begin{table*}[t]
\small
\caption{Controllability comparison of related methods on four key video elements: trajectory (orientation reported separately for clarity), action, subject and background. \ding{51} denotes standalone and accurate control, while \ding{55} indicates limited, inaccurate, or joint control.}
\label{tab:traj_orient_compare}
\vspace{-0.4cm}
\begin{center}
\begin{tabular}{lccccccc}
  \toprule
  \scalebox{1}{Class}   &  \scalebox{1}{\makecell{Example Methods}}  & \scalebox{1}{\makecell{Trajectory}} &  \scalebox{1}{\makecell{Orientation}} & \scalebox{1}{\makecell{Action}} & \scalebox{1}{\makecell{Subject}} & \scalebox{1}{\makecell{Background}}  \\ \midrule
  T2V/I2V Base Models & Wan-2.1~\cite{wang2025wan}, \etc & \ding{55}  & \ding{55}  & \ding{55} & \ding{55} (joint) & \ding{55} (joint) \\
  Image Animation & Animate Anyone~\cite{hu2024animate}, \etc & \ding{55} (2D) & \ding{55} (2D) & \ding{55} (2D) & \ding{55} (joint) & \ding{55} (joint)  \\
  \multirow{3}{*}{Motion Control}  & Tora~\cite{zhang2024tora} & \ding{55} (2D) & \ding{55}  & \ding{55}  & \ding{55} (joint)  & \ding{55} (joint)  \\
   & MotionCtrl~\cite{wang2024motionctrl} & \ding{55} (2D) & \ding{55}  & \ding{55}  & \ding{55} (text) & \ding{55} (text)  \\
   & 3DTrajMaster~\cite{fu20243dtrajmaster} & \ding{51} (3D) & \ding{51} (3D)  & \ding{55}  & \ding{55} (text) & \ding{55} (text) \\
  \midrule
  \multicolumn{2}{c}{\textbf{RealisMotion} (ours)}  & \ding{51} (3D) & \ding{51} (3D) & \ding{51} (3D) & \ding{51} (image) & \ding{51} (image)  \\
  \bottomrule
  \vspace{-1cm}
\end{tabular}
\end{center}
\centering
\end{table*}

\section{Related Work}
\subsection{Motion Acquisition}
To generate human motion, one can directly estimate human motion by motion capture systems, which are often prohibitively expensive. With advancements in human motion recovery techniques, extracting human motion from images or videos has become significantly simpler and more accessible~\cite{kanazawa2018end, goel2023humans, shin2024wham, wang2024tram, shen2024world, zhang2024rohm, yin2024whac}. These methods predominantly use learnable neural networks to directly predict the parametric human model parameters in SMPL~\cite{bogo2016keep, loper2023smpl} or SMPL-X~\cite{pavlakos2019expressive}. Most of them follow a multi-stage pipeline that consists of human bounding box tracking, 2D human keypoint detection, image feature extraction, camera relative rotation estimation and SMPL parameter regression. According to the difference of used coordinate systems, above methods can be roughly divided as camera-space~\cite{kanazawa2018end, goel2023humans, zhang2024rohm} and world-space~\cite{shin2024wham, shen2024world, wang2024tram, yin2024whac} methods. The former kind of method treats the camera as the origin and often fails to recover global motion due to accumulated translation and pose errors. In contrast, the latter kind of method defines a unified coordinate system without the impact of changing camera views, making it more suitable for subsequent motion editing. 

Another way for motion generation is training generative models based on captured human motion datasets~\cite{punnakkal2021babel, guo2022generating}. Given different guidance, such as action label~\cite{cervantes2022implicit}, audio~\cite{aristidou2022rhythm} and natural language~\cite{ahuja2019language2pose, tevet2022motionclip, tevet2022human, barquero2024seamless}, most methods choose conditional generative models to map from the conditioning domain to the motion domain. With significant advancements in diffusion models~\cite{sohl2015deep, ho2020denoising}, many methods start to train diffusion models for human motions conditioning on texts~\cite{tevet2022human, kim2023flame, shafir2023human, barquero2024seamless, zou2024parco}. For example, as one of the pioneering text-to-motion method, MDM~\cite{tevet2022human} adopts a transformer diffusion model for motion generation based on the CLIP text embedding.

\subsection{Motion-Guided Video Generation}
Similar to text-to-motion generation, diffusion-based models~\cite{blattmann2023stable, zhu2023denoising, yang2024cogvideox, liang2024movideo, kong2024hunyuanvideo, wang2025wan, chen2025lumosflow} have emerged as the current research mainstream for motion-guided video generation. As one of the pioneering methods, DisCo~\cite{wang2023disco} segments the foreground and background of the reference image, and then injects their VAE embeddings~\cite{esser2021taming} to the 2D UNet of Stable Diffusion~\cite{blattmann2023stable} by cross attention and ControlNet~\cite{zhang2023adding}, respectively. The 2D pose sequence is encoded and injected into the UNet by ControlNet as well. As another representative method, Animate Anyone~\cite{hu2024animate} upgrades the 2D UNet  to a 3D UNet for better video quality. It also proposes a symmetric ReferenceNet to extract reference features, which are merged into the main network via spatial attention. The feature of 2D pose sequence is concatenated with the noise input for motion guidance. Subsequent methods basically follow the designs of DisCo and Animate Anyone, with improvements on base models~\cite{zhang2024mimicmotion, lin2025omnihuman}, reference injection~\cite{xu2024magicanimate, wang2024unianimate, zhou2025realisdance, jiang2025vace}, motion guidance~\cite{zhu2024champ, tan2024animate, men2024mimo}, hand fidelity~\cite{zhou2024realisdance}, camera control~\cite{wang2024humanvid, shao2024human4dit}, object interaction~\cite{hu2025animate}, \etc. Some of them~\cite{zhu2024champ, zhou2025realisdance} have used the SMPL models, but their exploration is limited to the camera space. It is worth pointing out that most above methods are essentially image animation methods, without any modification on extracted motions from existing videos. Artifacts might arise when the motion (generally represented in rendered 2D image space) mismatches with the reference image.

In particular, 3DTrajMaster~\cite{fu20243dtrajmaster} attempts to control the object orientation and trajectory by representing them as the rotation-translation matrix, which is added with text embeddings to control video contents after cross attention. Since the non-rigid object motion is in fact defined by text prompts, it does not support complex and accurate motion control. Additionally, other techniques for modifying trajectories exist~\cite{yin2023dragnuwa, wang2024motionctrl, wu2024draganything}; however, the majority are limited to handling 2D rigid object movement and are not effective for intricate non-rigid human motion.

\begin{figure*}[ht]
\captionsetup{font=small}
\small
\begin{center}
\includegraphics[width=0.99\textwidth]{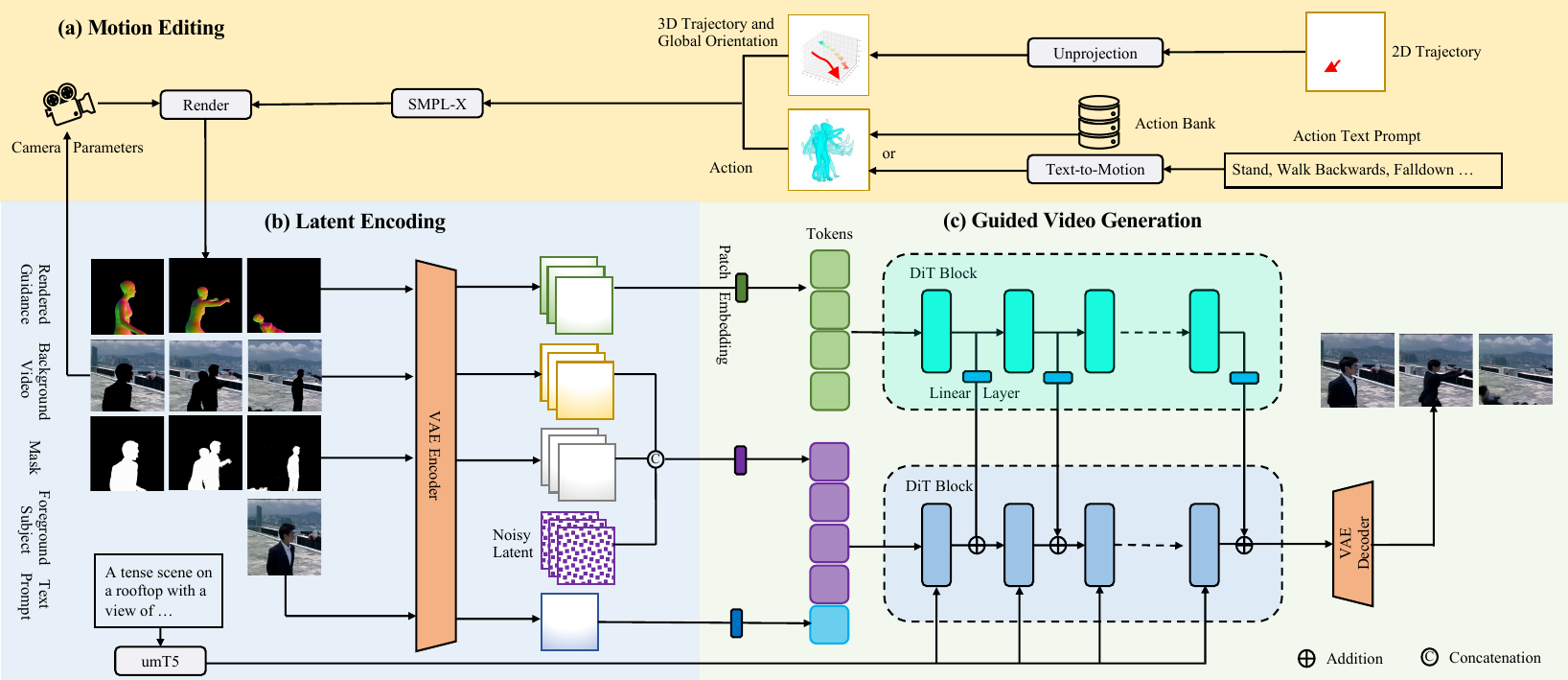}
\caption{The architecture of the proposed RealisMotion. It has two stages: 1) we first build a ground-aware 3D world coordinate system for the human motion, and conduct trajectory and action editing separately within the 3D space. 2) we then generate human videos conditional on the foreground subject image, background video and rendered motion guidance videos.}\label{fig:architecture}
\end{center}
\vspace{-0.3cm}
\end{figure*}

\section{Method} 
\subsection{Overall Pipeline} 
Given a reference human subject image $I$, a reference background video $V_{bgd}^{1:N}$, a sequence of target translation $T^{1:N}$ (also known as global trajectory), a sequence of target orientation $O^{1:N}$ and a sequence of target body pose $P^{1:N}$ (also referred to as human action), the goal in this paper is to generate a new video of the reference human moving in the background, following the defined motion (including $T^{1:N}$, $O^{1:N}$ and $P^{1:N}$). $N$ is the number of frames. 

To achieve the goal, we first match the motion with the background in Sec.~\ref{sub:motion}. Given the environment defined by the background video, the motion should follow the physical laws to ensure it appears reasonable and natural. Then, in Sec.~\ref{sec:motion-to-video}, we propose a motion-guided video generation model that supports separate subject, background and motion control. By this two-stage design, we combine the 3D physical prior with the learned video diffusion prior for generating highly realistic human motion videos. We solve the 3D-related problems, such as 3D trajectory control, 3D global orientation control, 3D action control, occlusion and foreshortening, in the 3D domain; and we solve the rest problems, such as object control, background control, detail authenticity, frame consistency, human-environment interaction, motion error repairing, in the video domain.

\subsection{Decoupled Motion Editing}
\label{sub:motion}

\subsubsection{Motion Representation}
We use the SMPL-X~\cite{pavlakos2019expressive} model for human body modelling in the low-level parametric space. It represents the human body as a function $\mathcal{M}(\gamma, \phi, \theta, \beta, \theta_h, \phi_f)$, which is parametrized by the global translation $\gamma\in\mathcal{R}^3$, global orientation $\phi\in\mathcal{R}^3$, body pose $\theta\in\mathcal{R}^{21\times3}$, body shape $\beta\in\mathcal{R}^{10}$, hand pose $\theta_h\in\mathcal{R}^{2\times15\times3}$ and facial expression $\phi_f$. After standard linear blend skinning and learned blend shape correction, the SMPL-X model outputs a 3D mesh representation with 10, 475 vertices. Hence, human motion could be well presented by a sequence of SMPL-X parameters.

To fit the motion into the background video, we need to make sure that both the motion and the environment in the background share the same 3D coordinate system: same coordinate origin, same axis direction and same coordinate scale. To avoid ambiguity, we build a world-grounded 3D coordinate system $(\overrightarrow{o}, \overrightarrow{x}, \overrightarrow{y}, \overrightarrow{z}, s)$ in the physical world without the impact of camera views in videos. More specifically, based on the human mesh recovery method GVHMR~\cite{shen2024world}, we define the coordinate system as follows: (a) the coordinate origin $\overrightarrow{o}$ is defined as the point where the human stands in the first frame of the video; (b) the $y$-axis $\overrightarrow{y}$ aligns with the gravity direction in the physical world; (c) we define the $x$-axis $\overrightarrow{x}$ and $z$-axis $\overrightarrow{z}$ as $\overrightarrow{x}=\overrightarrow{y} \times \overrightarrow{c}$ and $\overrightarrow{z}=\overrightarrow{x} \times \overrightarrow{y}$, respectively, where $\overrightarrow{c}$ is the camera view direction. In fact, it is difficult to align $\overrightarrow{x}$ and $\overrightarrow{z}$ for different $\overrightarrow{c}$, but we found that the x-z plane will always align with the ground plane given the definition of $\overrightarrow{o}$ and $\overrightarrow{y}$. Therefore, we can omit the mismatch of motion and environment in terms of $\overrightarrow{x}$ and $\overrightarrow{z}$, and rotate the 3D mesh with the rotation angle $\alpha$ between these two coordinate systems; (d) the coordinate scale $s$ is aligned with the physical distance, which means that a distance $d=1$ in the coordinate system means 1 meter in the physical world.

\subsubsection{Trajectory and Global Orientation Editing}
With the SMPL-X model, we can directly change its parameters $\gamma$ and $\phi$ to control the trajectory $\Gamma^{1:N}$ and global orientations $\Phi^{1:N}$, where $N$ is the length of points in the given trajectory. Since editing these 3D parameters manually frame-by-frame is labor-intensive, we propose to first obtain the 2D trajectory, and then derive the 3D trajectory and the corresponding orientations based on two reasonable assumptions: (a) the human moves on the ground; (b) the human faces the direction of movement. The 2D points can be easily obtained by dragging the cursor or by selecting a few key points and applying linear interpolation.

Formally, given a 2D point $\gamma^{n}_{2d}$ from the trajectory $\{\Gamma^{1}_{2d},\Gamma^{2}_{2d},...,\Gamma^{N}_{2d}\}$ on the image, we represent it as $\Gamma^{n}_{h}$ in the homogeneous 2D image coordinates and unproject it to the 3D camera space as
\begin{align}
    \Gamma^{n}_{c} = K^{-1}\Gamma^{n}_{h} \cdot d * f_2 / f_1
\end{align}
where $K$ and $f_1$ are the camera intrinsic matrix and focal length predicted by GVHMR. $d$ and $f_2$ are the depth and focal length estimated by Depth Pro~\cite{bochkovskii2024depth}. Here, we use $f_2 / f_1$ for calibration as GVHMR only predicts a fake focal length according to the image size, which might lead to inaccurate transformations during motion editing.

Then, we further transform the 3D point $\Gamma^{n}_{c}$ from the camera space to the defined world space as 
\begin{align}
    \Gamma^{n}_{w} = (\Gamma^{n}_{c}-T_{w2c}) R_{w2c}^{-1}
\end{align}
where $R_{w2c}$ and $T_{w2c}$ are the rotation matrix and translation vector from the world space to the camera space. $R_{w2c}$ and $T_{w2c}$ are calculated based on the rigid point registration~\cite{umeyama1991least} of 3D human points between the world space and camera space in the background video. 

Next, to make sure that the human moves with natural speed on the edited trajectory, we align the speed of the edited trajectory with the original speed. Otherwise, motion flaws such as feet sliding may occur when the feet move forward instead of maintaining static contact with the ground as would be expected in natural human motion. In detail, the alignment process starts with accumulating the total moving distance $\Delta^n$ from the first frame to the $n$-th frame as 
\begin{align}
\Delta^n = \sum_{i=2}^{n}\|\Gamma^{i}_{w}-\Gamma^{i-1}_{w}\|_1
\end{align}
where $\|\cdot\|_1$ means the $\mathcal{L}_1$ norm. When we fit $\Delta^n$ and the edited translation $\Gamma^n$ as a function $\Gamma^n=\mathcal{F}(\Delta^n)$ for $n=1,...,N$, we can obtain the aligned translation $\bar{\Gamma}^n$ as $\bar{\Gamma}^n=\mathcal{F}(\Delta'^n)$, where the original total moving distance $\Delta'^n$ is defined similarly to $\Delta^n$ for the original trajectory. 

After editing the trajectory, we edit the global orientation accordingly. For each frame $n$, we obtain the rotation angle $\Psi^n$ on the $x$-$z$ plane and derive the rotation matrix $R_n$ as
\begin{align}
\Psi^n=atan(\frac{z^{n}-z^{n-1}}{x^n-x^{n-1}}),
R^n = 
\begin{bmatrix}
cos(\Psi^n) & 0 & -sin(\Psi^n) \\
0 & 1 & 0 \\
sin(\Psi^n) & 0 & cos(\Psi^n)
\end{bmatrix}.
\end{align}
To change the human orientation, we found that directly modifying $\Phi_n$ leads to unnatural swinging movements. Therefore, we apply the trajectory and orientation transformations together on 3D human vertices $\mathcal{V}^n$ as 
\begin{align}
    \bar{\mathcal{V}}^n = (\Phi^n)^{-1}(\mathcal{V}^n - \Gamma^n)R^n + \bar{\Gamma}^n
\end{align}

Notably, due to the estimation errors, the edited human motion might suffer from feet floating or penetration to the ground. We shift vertices along the $y$-axis by subtracting the minimum $y$ value over a local temporal window to optimize foot contact. Besides, to improve motion consistency across frames, we also smooth the rotation angle in a sliding way during orientation editing.

\subsubsection{Body Pose and Hand Pose Editing}
For body pose and hand pose, we can directly copy them from existing SMPL-X parameters. Consequently, we can easily collect a motion bank from existing videos with extracted SMPL-X parameters. When we use the motion to generate new videos, we just need to edit the trajectory and orientation according to the background, while the body pose and hand pose are kept unchanged. This allows us to retrieve different actions, such as walking, running and swimming, with their original action styles, from the motion bank. For repetitive motions, one can cut a clip of motion and repeat it as needed. As for the editing of body pose and hand pose, it is out of the scope of this paper and the readers can refer to related research such as~\cite{agrawal2023pose, li2024unipose}.

In practice, the hand orientation $\Phi_h^n$ and hand pose $\Theta_h^n$ are estimated with an extra hand mesh recovery method HaMeR~\cite{pavlakos2024reconstructing}. It uses the parametric hand model MANO~\cite{romero2022embodied} and estimates the hand parameters in the camera space. To match the hand with the human body in the world space, a quick solution is to match the HaMeR hand vertices with the SMPL-X hand vertices using rigid point registration, but it might result in incorrect waist rotations when the hand pose is significantly different from the standard hand pose of SMPL-X. Hence, we match the hand orientation parameters between MANO and SMPL-X by first reversing the original SMPL-X hand orientation and then apply the MANO orientation after camera-world space transformation. This is formulated as
\begin{align}
    \bar{\Phi}_h^n=(\Omega^n)^{-1}(\Phi^n_{h}R_{w2c}^{-1})
\end{align}
where $\Omega^n$ is the hand orientation derived from the SMPL-X model using forward kinematics.

\subsubsection{2D Guidance Rendering}
Given the 3D human mesh representation, we render 2D depth maps, normal maps, and color maps to guide the video generation process. The same extrinsic and intrinsic camera parameters as the background video are used to ensure that the guidance maps and the target video are spatially aligned. In particular, the depth maps depict the distances from the camera to each pixel, while the normal maps contain the surface orientations of the meshes. Both of them provide critical geometric information for reconstructing the 3D structure of the human. Similar to RealisDance~\cite{zhou2024realisdance}, we generate color maps by assigning different colors to different vertices, which can provide semantic information for different parts of the human, and improves human consistency across different frames. We also refer to RealisDance for rendering the hand maps. One thing to note is that we need to mask the occluded hand by comparing the depths of human body and hand. In addition, after we transferring motion from one human to the reference human subject, we use the body shape parameters $\beta$ of the reference subject, which allows us to keep the same body shape such as height and figure. When transferring motion from adults to children, we add an extra shape parameter to interpolate between SMPL-X and SMIL-X templates~\cite{patel2021agora, hesse2018learning}.

\begin{table*}[t]
\small
\caption{Comparison of trajectory and global orientation control with existing methods on the proposed Trajectory100 dataset.}
\label{tab:traj_orient}
\vspace{-0.5cm}
\begin{center}
\begin{tabular}{lccccccc}
  \toprule
  \scalebox{1}{Method}   &  \scalebox{1}{\makecell{Translation Error (m)$\downarrow$}} & \scalebox{1}{\makecell{Rotation Error (deg) $\downarrow$}} & \scalebox{1}{PSNR$\uparrow$} & \scalebox{1}{SSIM$\uparrow$} & \scalebox{1}{LPIPS$\downarrow$}& \scalebox{1}{FID$\downarrow$}  & \scalebox{1}{FVD$\downarrow$} \\ \midrule
  Wan-2.1-I2V~\cite{wang2025wan} & 10.349 & 0.418 & 14.96 & 0.4763 & 0.3260 & 33.06 & 1421.87  \\
  Tora~\cite{zhang2024tora} & 5.667 & 0.355 & \underline{16.56} & \underline{0.5195} & 0.2501 & \underline{21.51} & 957.81\\
  RealisDance-DiT~\cite{zhou2025realisdance} & \underline{1.706} & \underline{0.167} & 16.17 & 0.4892 & \underline{0.2481} & {23.02}  & \underline{758.08}\\
  \midrule
  \textbf{RealisMotion} (ours) & \textbf{1.198} & \textbf{0.101} & \textbf{22.57} & \textbf{0.7664} & \textbf{0.0686} & \textbf{12.00} & \textbf{314.59}\\
  \bottomrule
  \vspace{-1cm}
\end{tabular}
\end{center}
\centering
\end{table*}

\subsection{Decomposed Human Video Generation}
\label{sec:motion-to-video}
We build our human video generation model based on the text-to-video model Wan-2.1~\cite{wang2025wan}, which achieves state-of-the-art performance on video generation. It compresses the video into the latent space with a spatio-temporal causal variational autoencoder (VAE)~\cite{esser2021taming} and employs full attention~\cite{vaswani2017transformer,peebles2023scalable} for spatio-temporal contextual modeling of video tokens. As shown in Fig.~\ref{fig:architecture}, we decompose the video into several key elements for flexible and separate control, including foreground subject, background video, motion guidance and text. 

\vspace{-0.4cm}
\paragraph{Subject Control}
To control the subject, we first compress the subject image as image tokens using the Wan-2.1 VAE. Then, the image tokens are concatenated with the video tokens for full attention. To discriminate between reference image and target video tokens, we treat the reference image as a sufficiently distant video frame in the target video (for example, the 80-th frame) and apply the corresponding rotary position embeddings (RoPE)~\cite{su2024roformer} on it. This leads to a sufficiently large distance between image and video tokens during attention, while keeping the spatial composition of the reference image. In addition, we found the generated human face might be blurry possibly due to the fact that the face often occupies a relatively small area of the whole image. To improve the face performance, we detect the face in the reference image and upscale it as an extra reference image input. An ID embedding module similar to the time embedding module in Wan-2.1 is proposed for distinguishing the reference subject image and face image.

\vspace{-0.4cm}
\paragraph{Background Control} 
To control the background of video, it is straightforward to compress the reference background as video tokens and then concatenate it with the target video tokens along the channel dimension, as the background video and the target video are supposed to be fully aligned. Typically, we obtain the background video with a human in it, especially in training. To avoid information leaking, we mask the foreground human in the background video with a mask. We also concatenate the mask with the video tokens along the channel dimension for helping the model identify the foreground area. In training, we additionally add random masks to background video to tackle with possible discrepancy between the target human area and masked foreground area during inference. 

\vspace{-0.4cm}
\paragraph{Motion Control} 
Given the rendered motion guidance videos, we encode them as visual tokens by VAE. Then, inspired by ControlNet~\cite{zhang2023adding}, we copy the transformer blocks $\mathcal{T}$ of Wan-2.1 as $\mathcal{T'}$ and extract motion features $\mathbf{c}$ from different blocks. Next, we add the motion features to the video features $\mathbf{x}$ at corresponding positions for controlling the video motion. This is formulated as
\begin{align}
    \mathbf{c}^{b+1} &= \mathcal{T'}^n(\mathbf{c}^b), ~~for~b=1,...,B\\
    \mathbf{x}^{b+1} &= \mathcal{T}^n(\mathbf{x}^b) + \mathcal{S}(\mathbf{c}^{b+1}), ~~for~b=1,...,B
\end{align}
where $b$ is the block index in $B$ blocks and $\mathcal{S}$ is a linear layer with zero initialization. To reduce model size and computation burden, we only use $B'$ blocks for motion feature extraction and add them to their neighboring blocks within a window size of $B/B'$. In other words, every $B/B'$ blocks share the same motion feature. 

\vspace{-0.4cm}
\paragraph{Text Control}
It seems that a combination of the subject image, background video and driving motion can define a video well. However, we found that providing the text is still important for improving the model performance, possibly due to two reasons. First, the Wan-2.1 model was trained for the text-to-video task. Removing the text-related modules or providing empty text might lead to significant domain gaps. Second, there are still some undefined elements in the video, such as the other side of the reference human subject, or the interaction of human and environment. Therefore, we keep the text modules and annotate the video with corresponding text prompts. Particularly, we avoid the cross attention between the reference image tokens and text tokens in text modules, as we observe a performance drop of reference ID preservation ability.

\vspace{-0.4cm}
\paragraph{The Image-to-Video Variant} We can seamlessly extend our model to the Wan-2.1 I2V (image-to-video) model, which additionally inputs the first frame of the video as a guidance. In this case, our model degenerates to be an image animation model when the reference subject and background are merged into a single image. It no longer supports separate subject-background customization, nor does it offer dynamic background control ability. We notice that there is a concurrent image animation work RealisDance-DiT~\cite{zhou2025realisdance}, which could be adopted as our I2V variant to prevent duplicate efforts.

\begin{figure}[t]
\captionsetup{font=small}
\small
\begin{center}
\includegraphics[width=0.45\textwidth]{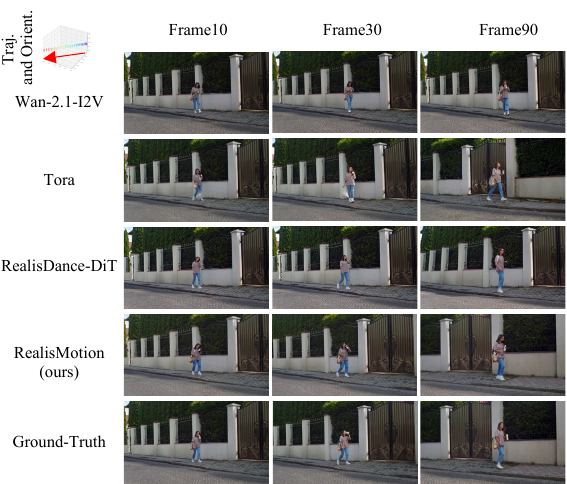}
\vspace{-0.1cm}
\caption{Visual comparison of different methods on trajectory and global orientation control. More visual results are provided in the supplementary.}
\label{fig:traj_orient}
\end{center}
\vspace{-0.4cm}
\end{figure}

\section{Experiments}
\subsection{Experimental Setup}
Based on the Wan-2.1 14B model, we finetune our model on an internal dataset that comprises approximately 3,300 hours of multi-resolution human video content. The details are provided in the supplementary due to page limit. For evaluation, we compare our methods in several aspects. For trajectory and global orientation control, we compare the translation error and rotation error defined by MotionCtrl~\cite{wang2024motionctrl}, and also report video quality metrics including PSNR, SSIM, LPIPS~\cite{zhang2018lpips}, FID~\cite{heusel2017gans} and FVD~\cite{unterthiner2019fvd}. For action control, we mainly compare the video metrics with existing image animation methods.

\subsection{Comparison with Existing Methods}
\subsubsection{Trajectory and Global Orientation Control}
To assess trajectory and global orientation control capabilities, we created a 100-video evaluation dataset with distinct movement paths, named Trajectory100. We compare our approach against the Wan-2.1 base model~\cite{wang2025wan}, the trajectory-focused method Tora~\cite{zhang2024tora}, and the image animation method RealisDance-DiT~\cite{zhou2025realisdance}. As illustrated in Table~\ref{tab:traj_orient}, our proposed RealisMotion outperforms all models in each metric. The lowest translation and rotation errors demonstrate superior trajectory and global orientation control, while additional metrics confirm that our generated videos also offer the highest visual quality. Fig.~\ref{fig:traj_orient} shows that although Tora and RealisDance-DiT can control human trajectories in the 2D camera space to some extent, their outputs do not accurately represent physical positions within the environment. Furthermore, related methods like MotionCtrl~\cite{wang2024motionctrl} and 3DTrajMaster~\cite{fu20243dtrajmaster} are excluded since their video backgrounds and objects are specified by text prompts, making quantitative evaluation on Trajectory100 difficult. A detailed comparison of controllability is available in Table~\ref{tab:traj_orient_compare}.

\begin{table}[t]
\centering
\small
\caption{Comparison of action control with existing methods on the RealisDance-Val~\cite{zhou2025realisdance}.}
\vspace{-0.2cm}
\tabcolsep=0.5mm
\begin{tabular}{lccccc}
    \toprule
    Method & PSNR$\uparrow$ & SSIM$\uparrow$ & LPIPS$\downarrow$& FID$\downarrow$ & FVD$\downarrow$  \\
    \midrule
    Animate-X~\cite{tan2024animate} & 16.29  & 0.5893  &  0.2664 & 36.50 & 2376.66\\
    ControlNeXt~\cite{peng2024controlnext} & 15.66 & 0.5762  &  0.2776 & 40.38 & 2412.52\\
    MimicMotion~\cite{zhang2024mimicmotion} & {17.20} & {0.6029}  & {0.2457} & 43.51 & 2283.93 \\
    MooreAA~\cite{hu2024animate} & 16.08  & 0.5546  &  0.2488 & 37.92 & 2446.50 \\
    MusePose~\cite{tong2024musepose} & \underline{17.29} & \underline{0.6080}  &  0.2276 & 44.66 & 2809.02 \\
    RealisDance-DiT~\cite{zhou2025realisdance} &  {17.22} & 0.5919 & \underline{0.2050} & \underline{26.18} & \underline{1576.66} \\
    \midrule
    \textbf{RealisMotion} (ours) & \textbf{20.34} & \textbf{0.7224} & \textbf{0.0998} & \textbf{20.67} & \textbf{1000.98}\\
    \bottomrule
\end{tabular}
\vspace{-0.3cm}
\label{tab:action_subject_background}
\end{table}

\subsubsection{Action Control}
We evaluate the action control performance of various methods using the image animation benchmark dataset RealisDance-Val~\cite{zhou2025realisdance}. As presented in Table~\ref{tab:action_subject_background}, RealisMotion significantly surpasses existing methods across all five metrics, demonstrating its robust action control capabilities. The qualitative results, depicted in Fig.~\ref{fig:action}, reveal that our approach produces clear, visually appealing videos with accurate actions, whereas the comparative methods often result in unnatural, distorted human figures.

\subsubsection{Subject and Background Control}
\begin{figure}[t]
\captionsetup{font=small}
\small
\begin{center}
\includegraphics[width=0.45\textwidth]{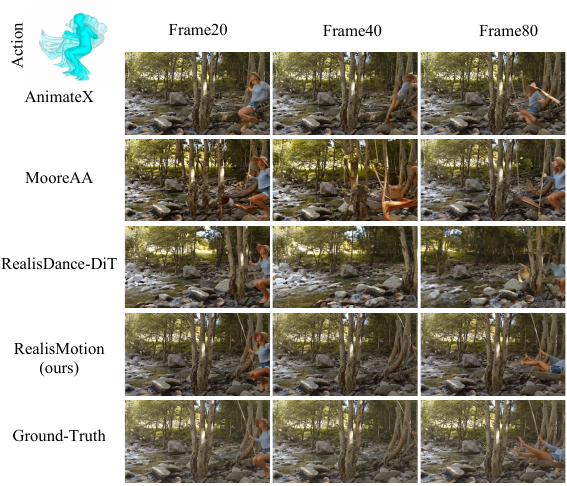}
\caption{Visual comparison of different methods on action control. More visual results are provided in the supplementary.}
\label{fig:action}
\end{center}
\vspace{-0.3cm}
\end{figure}

\begin{figure}[t]
\captionsetup{font=small}
\small
\begin{center}
\includegraphics[width=0.45\textwidth]{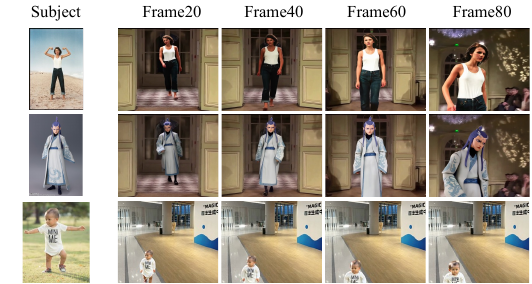}
\caption{Visual results of subject control. More visual results are provided in the supplementary.}
\label{fig:subject}
\end{center}
\vspace{-0.3cm}
\end{figure}

As depicted in Fig.~\ref{fig:teaser} and Fig.~\ref{fig:subject}, our approach allows for arbitrary subject customization and movement within existing background videos by referring to a reference image. Although our model has been mainly trained on adult human videos, it demonstrates strong generalization capabilities to previously unseen animation characters and children. In terms of background control, the effectiveness of our approach is illustrated in the last two rows of Fig.~\ref{fig:traj_orient} and Fig.~\ref{fig:action}, wherein it consistently preserves background continuity, a feature not observed in the comparative methods. Note that the recent Animate Anyone 2~\cite{hu2025animate} is not compared here as it is not open-sourced.

\subsection{Ablation Study}
We conduct ablation study on Trajectory100. The accompanying visual comparison and additional ablation studies are provided in the supplementary.

\vspace{-0.4cm}
\paragraph{Focal Length Calibration}
To mitigate the adverse effects of inaccurate focal length, we calibrate the focal length. As demonstrated in Table~\ref{tab:ablation_study}, the PSNR decreases from 22.57dB to 21.52dB when calibration is absent. Visual examples in the supplementary material reveal that, without calibration, the human size may appear inconsistent with the surrounding environment, thereby contravening physical commonsense.

\vspace{-0.4cm}
\paragraph{Body-Hand Matching}
Given that the human body and hands are predicted using different methods and within different spaces, we align the hands with the body to achieve more precise hand pose control. In the absence of this alignment, the default hand pose is used, resulting in a decrease in PSNR to 22.34dB.

\vspace{-0.4cm}
\paragraph{Text Prompt} 
Since the foreground, background, and motion effectively define a video, we attempt to remove the text module to reduce computational demands and simplify the inference process. However, as indicated in Table~\ref{tab:ablation_study}, this leads to a performance drop in video quality. The visual results provided in the supplementary reveal that the resulting videos tend to generate incorrect details.

\vspace{-0.4cm}
\paragraph{Shifted RoPE for Reference Subject Image}
We propose to shift the RoPE to differentiate between the reference image and the target video. Without this design, the PSNR decreases to 22.13dB. The visual results in the supplementary material show that the first frame deteriorates significantly, likely because the absence of RoPE on the reference frames actually causes the reference frame to be treated as the first frame.

\vspace{-0.4cm}
\paragraph{Extra Face Image for Reference}
With an additional face image input, the PSNR improves from 22.36dB to 22.57dB. This enhancement is further corroborated by the visual comparisons provided in the supplementary.

\vspace{-0.4cm}
\paragraph{Random Masking On Background}
We randomly apply masking to the background to address the mismatch between background and motion during inference. As illustrated in the supplementary, the absence of random masking can lead to the generation of two human figures: one in the original human region and another in the new motion region, resulting in significant performance drops, as indicated in Table~\ref{tab:ablation_study}.

\begin{table}[t]
\centering
\small
\tabcolsep=3mm
\caption{Ablation Study on different designs. The accompanying visual results are provided in the supplementary.}
\begin{tabular}{lcc}
    \toprule
    Ablation Study (w/o) & PSNR$\uparrow$ & LPIPS$\downarrow$ \\
    \midrule
    Focal Length Calibration & 21.52 & 0.1043 \\
    Body Hand Matching & 22.34 & 0.0694\\
    Text Prompt & 22.12 & 0.0793 \\
    Extra Face Input & 22.36 & 0.0701 \\
    Shifted RoPE & 22.13 & 0.0752\\
    Random Masking & 21.88 & 0.0951\\
    \midrule
    \textbf{RealisMotion} (ours) & \textbf{22.57} & \textbf{0.0686}\\
    \bottomrule
\end{tabular}
\label{tab:ablation_study}
\end{table}

\section{Conclusions}
In this paper, we present RealisMotion, a decomposed human motion control and video generation framework. It constructs a ground-aware 3D world coordinate system that enables straightforward, realistic trajectory and action editing in the 3D space. Using the rendered motion guidance, RealisMotion synthesizes videos with independent control over foreground subject, background, trajectory, and action. Extensive experiments demonstrate state-of-the-art video quality and superior controllability across these elements.

\vspace{-0.4cm}
\paragraph{Limitation and Future Work} Currently, our method has limited sensitivity to the environment's 3D structure and can sometimes produce foreground–background lighting inconsistencies. We leave these challenges for our future work.

{\small
\bibliographystyle{ieee_fullname}
\bibliography{main.bib}
}

\end{document}